\documentclass[10pt]{article}
\usepackage[affil-it]{authblk}
\font\yyfont=cmr12 at 6.467pt
\font\myfont=cmr12 at 14.54pt
\usepackage[margin=1in]{geometry}
\usepackage{graphicx}
\graphicspath{ {/users/ck/desktop/} }
\usepackage{sectsty}
\usepackage{amssymb}
\usepackage{amsmath}
\usepackage{esvect}
 
\begin{document}

\title{{\myfont Optimization of Ensemble Supervised Learning Algorithms for Increased Sensitivity, Specificity,
and AUC of Population-Based Colorectal Cancer Screenings}}

\author[1]{Anirudh Kamath}
\author[2]{Raj Ramnani}
\author[3]{Jay Shenoy}
\author[4]{Aditya Singh}
\author[5]{Ayush Vyas} 
\affil[1]{Northeastern University}
\affil[2]{Yale University}
\affil[3]{University of California--Berkeley}
\affil[4]{The Wharton School, University of Pennsylvania}
\affil[5]{The Harker School}
\date{July 6, 2017}
\maketitle
\thispagestyle{empty}

\abstract{Over 150,000 new people in the United States are diagnosed with colorectal cancer each year.
Nearly a third die from it (American Cancer Society). The only approved noninvasive
diagnosis tools currently involve fecal blood count tests (FOBTs) or stool DNA tests. Fecal blood count
tests take only five minutes and are available over the counter for as low as \$15. They are highly
specific, yet not nearly as sensitive, yielding a high percentage (25\%) of false negatives (Colon
Cancer Alliance). Moreover, FOBT results are far too generalized, meaning that a positive result could mean much more than just colorectal cancer, and could just as easily mean hemorrhoids, anal fissure, proctitis, Crohn's disease, diverticulosis, ulcerative colitis, rectal ulcer, rectal prolapse, ischemic colitis, angiodysplasia, rectal trauma, proctitis from radiation therapy, and others. Stool DNA tests, the modern benchmark for CRC screening, have a much
higher sensitivity and specificity, but also cost \$600, take two weeks to process, and are not for high-risk individuals or people with a history of polyps. To
yield a cheap and effective CRC screening alternative, a unique ensemble-based classification algorithm
is put in place that considers the FIT result, BMI, smoking history, and diabetic status of patients. This method is tested under ten-fold cross validation to have a .95 AUC, 92\% specificity, 89\% sensitivity, .88 F1, and 90\% precision.
Once clinically validated, this test promises to be cheaper, faster, and potentially more
accurate when compared to a stool DNA test.}\\

{\yyfont This work is licensed under the Creative Commons Attribution-NonCommercial-ShareAlike 4.0
International License. To view a copy of this license, visit
http://www.creativecommons.org/licenses/by-nc-sa/4.0/.}
\begin{figure}[h]
\centering
\includegraphics[scale = 0.5]{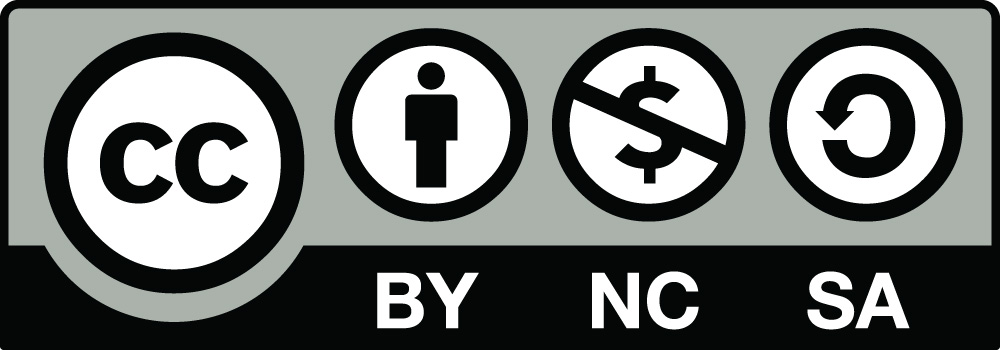}
\end{figure}
\section{Overview}

Colorectal cancer (CRC) is the third most common cancer and the second leading cause
of cancer-related deaths in the United States. The American Cancer Society predicts
approximately 135,000 new cases of CRC in 2017 and 50,000 deaths just in the United States.
There is a strong inverse association with CRC screening and CRC incidence and mortality, yet
more than one in three eligible Americans are not up-to-date with the recommended screening
(Burke and Mankaney, 2017).

\begin{figure}[h]
\caption{Source: Cancer Treatment Centers of America}
\centering
\includegraphics[scale = 0.4]{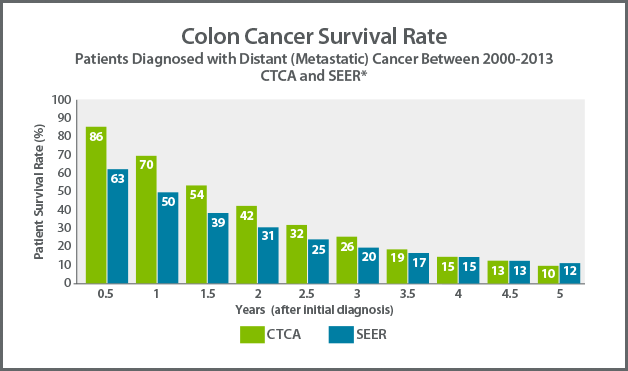}
\end{figure}

\section{Problem}
Once detected, a colonoscopy can typically remove most polyps or small cancers, but as
stated earlier, screening itself is a major problem. Current noninvasive tests are most commonly
fecal occult blood count tests or stool DNA tests.\\

{\bfseries Fecal Occult Blood Count Tests}\\\\
These tests, most commonly seen as guaiac fecal occult blood tests (gFOBT) or fecal
immunochemical tests (FIT) inject biochemicals into the stool to account for occult, or hidden, blood. Occult blood in the stool is typically a biomarker for bleeding polyps, which very often
grow into carcinomas. They are currently what most doctors use for CRC screening due to their
extremely high patient compliance from their accessibility and ease of use. The test is available
over-the-counter, takes five minutes, and is completely done at home. There is no further work
needed once the patient receives his/her results. Moreover, these tests are highly specific,
meaning the people it falsely marks positive is less than 5\%. The problem arises with its
use cases. A positive FIT result is simply indicative of abnormal gastrointestinal activity and not necessarily the presence of CRC. For example, causes of blood in the stool just as easily indicate the presence of hemorrhoids, anal fissure, proctitis, Crohn's disease, diverticulosis, ulcerative colitis, rectal prolapse, ischemic colitis, angiodysplasia, rectal trauma, proctitis from radiation therapy, and others. Therefore, the FIT is not the most reliable test to screen for just CRC.\\

{\bfseries Stool DNA Testing}\\\\
At this time of writing, the only such test is Cologuard, developed by Exact Sciences.
This test is brand new in the market, but it has shown very promising results. This method
analyzes the genes expressed in fecal matter and finds the ones associated with CRC. As a result,
its complicated analysis requires a lab test as opposed to a home test like occult blood testing,
which can take up to two weeks to get a result instead of the five-minute occult blood test.
Moreover, the test is not designed for high-risk individuals, namely people who have had a
history of polyps/cancer or for people for whom CRC runs in the family. On top of that, it comes
at a cost upwards of \$600, with availability requiring a prescription. This is quite easily
understood because these conditions alter stool DNA that may lead to inaccurate results.
Regardless, its specificity is 89\%, while sensitivity is 92\% (Colon Cancer Alliance). These
results are stunning, given the comparison to the occult blood testing. However, the price,
accessibility, and restrictions on who can use it have yielded hesitation in both doctors and
patients. 

\section[yfont]{Solution}
CounteractIO calculates predisposition to CRC through inputs such as age, BMI, diabetes, and smoking habits. When combined with results of fecal occult blood testing, ensemble learning methods are put into place in order to boost specificity/sensitivity such that a patient can have the accessibility,
ease, and quickness of an occult blood test with the accuracy of a stool DNA test.  \\\\

{\bfseries Process}\\\\
The process is rather straightforward. The patient takes an occult blood test, then opens
the CounteractIO app. The app will ask for the patient's age, BMI, presence of diabetes, and smoking habits and use these inputs to come up with a probability result of positive or negative for CRC. Since the app does not collect personal identifying information such as name or even date of birth, it should not be in
violation of any HIPAA regulations. The entirety of the system was developed using data from
Dr. Schloss' lab at the University of Michigan (Baxter et al., 2016).\\\\

{\bfseries Ensemble Learning}\\\\
The following definitions are used to present the underlying concepts of ensemble
methods according to classic literature (Kuncheva, 2004). \\\\
1. Let $ \Omega = \{\omega_1, \omega_2 ... \omega_M\} $ be a set of class labels. Then, a function $ D: \mathbb{R}^n \xrightarrow {} \Omega $ is called a
classifier, while a vector $\vv{\chi} = ({\chi}^1 + {\chi}^2 + ... {\chi}^n)  \in  \mathbb{R}^n$
is called a feature vector.\\\\
2. Let ${h}_1, {h}_2...{h}_M, {h}_i : \mathbb{R}^n \xrightarrow {} \mathbb{R}, i = 1...M$ be so-called discriminator functions 
corresponding to the class labels $\omega_1, \omega_2 ... \omega_M$, respectively. Then, the classifier $D$
belonging to these discriminator functions is defined by $D(\vv{\chi}) = \omega_j* \Leftrightarrow {h}_j*(\vv{\chi}) = \max_{1\leq j<M} ({h}_j(\vv{\chi}))$ for all $\chi \in \mathbb{R}^n$\\\\
3. Let $ {D}_1, {D}_2 ... {D}_L $ be classifiers. Then, the majority voting ensemble classifier ${D}_{maj} : \mathbb{R}^n \xrightarrow {} \Omega$ formed from these classifiers is defined as
${D}_{maj}(\vv{\chi})  = \omega_i* \Leftrightarrow | \{j : {D}_j (\vv{\chi}) = \omega_i, j = 1...M\}|$.\\\\

{\bfseries Inputs}\\\\
The inputs, represented as feature vectors, are as such:
\\\\
\begin{enumerate}
\item{$\chi_1$--FIT result: this is the result of the occult blood test. Blood in the stool is
presents a high risk of CRC.}
\item{$\chi_2$--BMI: a unique weight/height ratio. Higher BMIs put a patient at higher risk of
CRC.}
\item{$\chi_3$--Age: older people are more likely to get CRC.}
\item{$\chi_4$--Diabetes: Presence of diabetes imposes a risk of CRC}
\item{$\chi_5$--Smoking: Patients who smoke are much more likely to get CRC than those who
do not.}\\\\
\end{enumerate}

{\bfseries Classifiers}\\\\
To select the best classifiers for classification, quite a few well-known classifiers were
trained on the dataset. Classifiers were selected using backward search methods investigated in
(Ruta and Gabrys, 2005) for a fixed set of classifiers. A backward search starts out with all
classifiers as part of the ensemble, but as the performance of the ensemble increases, classifiers
are then removed, thereby resulting in the optimal ensemble. The end result ensemble is 
composed of extreme gradient boosted trees, logistic regression, random forests, decision trees,
an artificial neural network, and a support vector machine. \\

Hyperparameters were adjusted to adjust for a small dataset with multiple inputs. Most
notably, the artificial neural network has an L-BFGS solver to account for a small dataset and
inter-related inputs in combination with a hyperbolic tangent activation function. The logistic
regression classifier has a liblinear solver for optimization of small datasets. The support vector
machine has a linear kernel. \\\\

{\bfseries Metrics}\\\\
Metrics were calculated as follows after ten-fold cross validation: \\\\
1. Precision: $\frac{TP}{TP+FP}$\\\\
2. Sensitivity/Recall: $\frac{TP}{TP+FN}$\\\\
3. Specificity:  $\frac{TN}{TN+FP}$\\\\
4. F-Score:  $\frac{2TP}{2TP+FN+FP}$\\\\
5. AUC: $\int_\infty^{-\infty} TPR(T) (-FPR'(T))dT$\\

\begin{center}
 \begin{tabular}{c c} 
 \textit{Classifier} &  \textit{Scores} \\ [0.5ex] 
 \hline
 eXtreme Gradient Boosted Trees & Precision: .89  \\
   & Sensitivity: .86\\
   & AUC: .95\\
   & Specificity: .91\\\\
Logistic Regression & Precision: .89\\
   & Sensitivity: .86\\
   & AUC: .94\\
   & Specificity: .92\\\\
Random Forest & Precision: .92\\
    & Sensitivity: .89\\
    & AUC: .93\\
    &Specificity: .91\\\\   
Support Vector Machine & Precision: .89\\
    & Sensitivity: .86\\
    & AUC: .95\\
    & Specificity: .92\\\\
Artificial Neural Network & Precision: .90\\
    & Sensitivity: .89\\
    & AUC: .95\\
    & Specificity: .89 \\\\
Majority Vote & Precision: .90\\
    & Sensitivity: .89\\
    & AUC: .95\\
    & Specificity: .92\\
    & F1: .88\\[1ex] 
\end{tabular}
\end{center}
These results were then extrapolated onto a test set and plotted on a receiver  operating 
characteristic (ROC) curve. The results are below:

\begin{figure}[h]
\hspace*{-0.1cm}
\centering
\includegraphics[scale = 0.3]{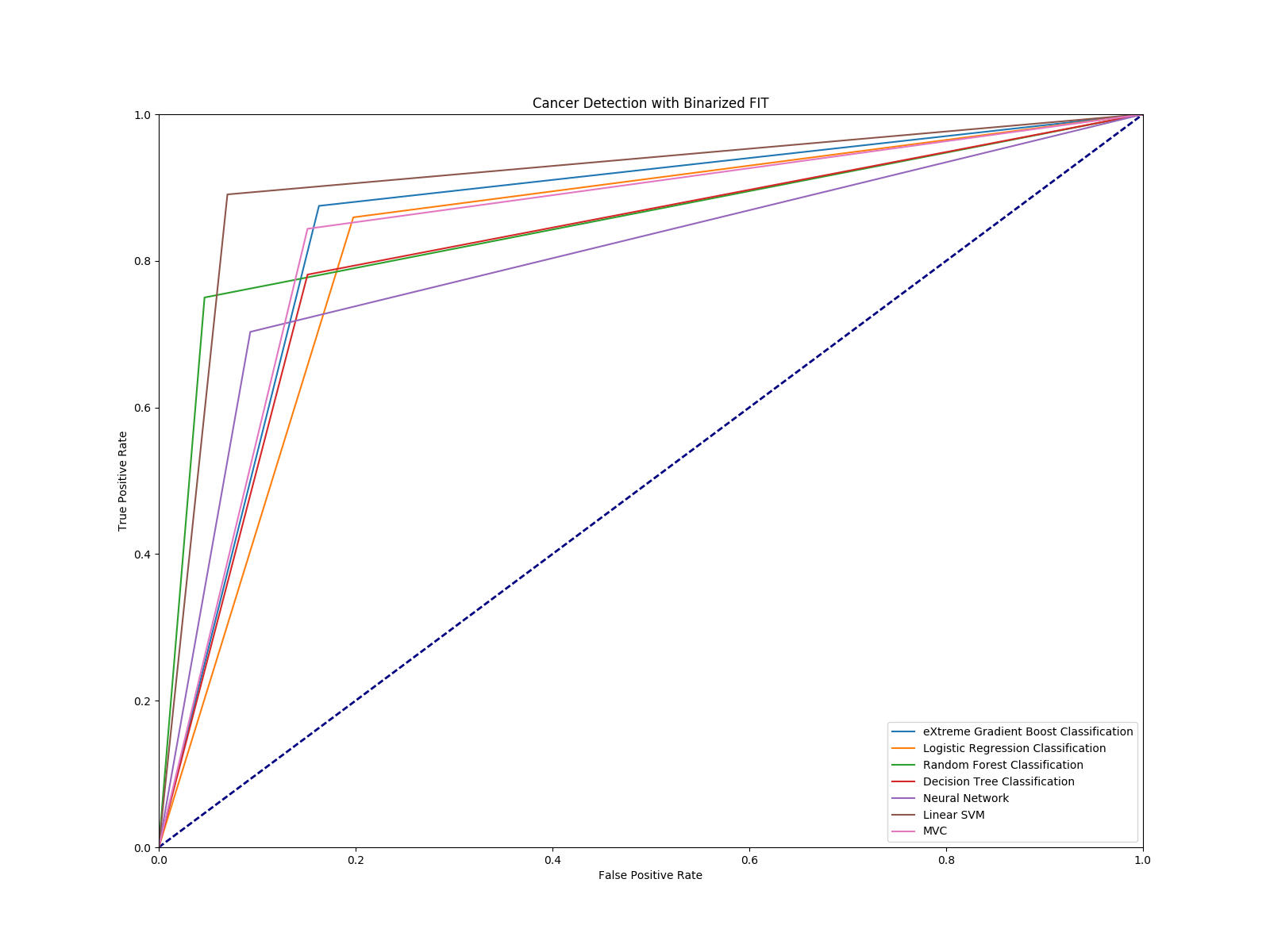}
\end{figure}
\begin{center}
{\bfseries Comparison with existing methods}
\begin{tabular}{c c} 
 \textit{Method} &  \textit{Statistics} \\ [0.5ex] 
 \hline
Fecal Occult Blood Tests & Specificity: .96  \\
   & Sensitivity: .74\\
   & Cost: \$15\\
   & Time: 5 minutes\\\\
CT Colonography & Specificity: .88  \\
   & Sensitivity: .84\\
   & Cost: \$439\\
   & Time: 30 minutes\\\\
Stool DNA Test & Specificity: .89\\
   & Sensitivity: .92\\
   & Cost: \$600\\
   & Time: 2 weeks\\\\
CounteractIO & Specificity: .92\\
    & Sensitivity: .89\\
    & Cost: \$10\\
    & Time: 5 minutes\\[1ex] 
\end{tabular}
\end{center}

\section{Shortcomings}
This has been tested for patients in the United States and Canada. Clinical research
locations include the MD Anderson Cancer Center, the University of Michigan in Ann Arbor, the Mayo Clinic, and the Dana Farber Cancer Institute, thereby representing people from all over the country. However, the methodology described in this paper has not been tested outside of the United States/Canada. This is significant because people outside of the States may show different biomarkers of CRC as a result of a different cultural upbringing. We are working on getting a larger and more diverse sample set to further
validate our results.

\section{Conclusion}
CounteractIO fills a void that has a great need in the current world of CRC
detection. Current tests are not accessible, accurate, quick, as well as noninvasive. The fact that
CounteractIO is able to fit all four of those criteria makes it highly desirable to patients in the
gastroenterology and oncology fields.

\section{References}
{[1]} Baxter, NT., Ruffin MT., Rogers M.A., Schloss, P.D. Microbiota-based model improves the
sensitivity of fecal immunochemical test for detecting colonic lesions. Genome Med.
2016. \\\\
{[2]} Burke, C., Mankaney, G., 2017. Colorectal Neoplasia. Cleveland Clinic.\\\\
{[3]} Colon Cancer Survival Statistics | CTCA. (0001, January 01). \\\\
{[4] }Key Statistics for Colorectal Cancer. (n.d.). American Cancer Society\\\\
{[5]} Kuncheva, L. I., 2004. Combining Pattern Classifiers. Methods and Algorithms. Wiley.\\\\
{[6]} Ruta, D., Gabrys, B., 2005. Classifier selection for majority voting. Information Fusion 6 (1), 63 -- 81\\\\
{[7] }Screening Methods - Colon Cancer Alliance - Prevention, Research, Patient Support. (n.d.).

\end{document}